\documentclass[10pt,twocolumn,letterpaper]{article}

\usepackage{cvpr}              

\usepackage{booktabs}
\usepackage{graphicx}
\usepackage{amsmath}
\usepackage{amssymb}
\usepackage{booktabs}

\usepackage{multirow}
\usepackage{bbding}

\usepackage{multirow}
\usepackage{graphicx, amsmath, amssymb, caption, subcaption, multirow, overpic, textpos}
\usepackage{pifont}
\usepackage{adjustbox}
\newcommand{\cmark}{\ding{51}}
\newcommand{\xmark}{\ding{55}}
\usepackage[table]{xcolor}
\usepackage[british, english, american]{babel}
\usepackage[pagebackref=false, breaklinks=true, letterpaper=true, colorlinks,citecolor=green, linkcolor=red, bookmarks=false]{hyperref}
\newlength\savewidth

\renewcommand{\paragraph}[1]{\vspace{1.25mm}\noindent\textbf{#1}}

\newcolumntype{x}[1]{>{\centering\arraybackslash}p{#1pt}}
\newcolumntype{y}[1]{>{\raggedright\arraybackslash}p{#1pt}}
\newcolumntype{z}[1]{>{\raggedleft\arraybackslash}p{#1pt}}

\def\cvprPaperID{1284} 
\def\confName{CVPR}
\def\confYear{2022}

\begin{document}

\title{Semantic-Aware Generation for Self-Supervised Visual Representation Learning}

\author{Yunjie Tian\textsuperscript{1}, Lingxi Xie\textsuperscript{2}, Xiaopeng Zhang\textsuperscript{2}, Jiemin Fang\textsuperscript{3}, Haohang Xu\textsuperscript{4}, Wei Huang\textsuperscript{1},\\
Jianbin Jiao\textsuperscript{1}, Qi Tian\textsuperscript{2}, Qixiang Ye\textsuperscript{1}\\
\textsuperscript{1}University of Chinese Academy of Science\quad \textsuperscript{2}Huawei Inc.,\\
\textsuperscript{3}Huazhong University of Science and Technology\quad\textsuperscript{4}Shanghai Jiao Tong University\\
{\tt\small tianyunjie19@mails.ucas.ac.cn}\quad{\tt\small 198808xc@gmail.com}\quad{\tt\small zxphistory@gmail.com}\\
{\tt\small jaminfong@hust.edu.cn}\quad{\tt\small xuhaohang@sjtu.edu.cn}\quad{\tt\small huangwei19@mails.ucas.ac.cn}\\
{\tt\small jiaojb@ucas.ac.cn}\quad{\tt\small tian.qi1@huawei.com}\quad{\tt\small qxye@ucas.ac.cn}
}
\maketitle

\begin{abstract}
In this paper, we propose a self-supervised visual representation learning approach which involves both generative and discriminative proxies, where we focus on the former part by requiring the target network to recover the original image based on the mid-level features. Different from prior work that mostly focuses on pixel-level similarity between the original and generated images, we advocate for \textbf{S}emantic-\textbf{a}ware \textbf{Ge}neration (SaGe\footnote{The code is at \textsf{https://github.com/sunsmarterjie/SaGe}}) to facilitate richer semantics rather than details to be preserved in the generated image. The core idea of implementing SaGe is to use an evaluator, a deep network that is pre-trained without labels, for extracting semantic-aware features. SaGe complements the target network with view-specific features and thus alleviates the semantic degradation brought by intensive data augmentations. We execute SaGe on ImageNet-1K and evaluate the pre-trained models on five downstream tasks including nearest neighbor test, linear classification, and fine-scaled image recognition, demonstrating its ability to learn stronger visual representations.
\end{abstract}

\section{Introduction}
\label{sec:intro}

Self-supervised visual representation learning has attracted increasing attentions of the community, arguably due to its potentials of extracting general and transferable features that apply to various downstream tasks. Compared to supervised learning where manual annotations are naturally used as learning objectives, the key of self-supervised learning is to design some type of proxies so that extracted features satisfy annotation-free constraints. In the past years, the community has witnessed the evolution from geometry-based proxies (\textit{e.g.}, solving jigsaw puzzles~\cite{jigsaw} and predicting rotations~\cite{rotation}) to contrastive~\cite{chen2020simple, he2020momentum} or predictive~\cite{grill2020bootstrap,bao2021beit} proxies. Despite the substantial progress in terms of downstream performance, we note that the above approaches mostly focused on the discriminative ability yet somewhat undervalued the generative ability -- as a reference, generation-based approaches~\cite{inpainting,aet} reported inferior performance in downstream tasks.

In this paper, we construct a self-supervised learning framework that requires the target network to gain the abilities of \textbf{both} discrimination and generation. Specifically, the framework consists of three parts, (i) an encoder (\textit{i.e.}, the target network that is transferred to downstream tasks) that extracts visual features into a compact vector; (ii) a decoder (\textit{i.e.}, a complementary network) that tries to recover the original image based on the compact vector; and (iii) an evaluator that measures the quality of the encoder-decoder system. The evaluator itself outputs two scores, where a discrimination score is computed by feeding the compact vector into either contrastive or predictive learning, and a generation score is obtained from a standalone module that takes both the original and recovered images as inputs.


\begin{table}[]
\fontsize{8.0}{10.5}\selectfont
\centering
\setlength{\tabcolsep}{0.08cm}
\begin{tabular}{x{100}|x{48}|x{32}|x{32}}
\toprule
\multirow{3}{*}{Method} & discrimination & \multicolumn{2}{c}{generation} \\
\cline{2-4}  &\multirow{2}{*}{\stackbox[c][c]{instance\\level}}    & \multirow{2}{*}{\stackbox[c][c]{pixel\\level}}    & \multirow{2}{*}{\stackbox[c][c]{semantic\\level}} \\
&&&\\
\hline
MoCo~\cite{he2020momentum}, SimCLR~\cite{chen2020simple}, BYOL~\cite{grill2020bootstrap}, SwAV~\cite{caron2020unsupervised}, \textit{etc.} & \multirow{2}{*}{\cmark} &\multirow{2}{*}{} &\multirow{2}{*}{}\\
\hline
Auto-encoder~\cite{hinton2006reducing}, inpainting~\cite{inpainting}, AET~\cite{aet}, Colorization~\cite{colorization}, \textit{etc.} &\multirow{3}{*}{}   &\multirow{3}{*}{\cmark} & \\
\hline
PCRL~\cite{medical_reconstruct}, RCL~\cite{making_yonglong}, GenRep~\cite{generative_multiview}, \textit{etc.}
 &\multirow{2}{*}{\cmark} &\multirow{2}{*}{\cmark} & \\
\hline
\textbf{SaGe} (ours) & \cmark & \cmark & \cmark \\
\bottomrule
\end{tabular}
\caption{A comparison between the proposed framework, \textbf{SaGe}, and prior self-supervised learning approaches. SaGe not only takes both discrimination and generation into consideration, but also enhances the generation branch with semantic awareness. }
\label{tab:introduction}
\end{table}

Based on the above framework, we offer a new insight to the aforementioned phenomenon that discrimination outperforms generation. The secret lies in how the generation score is computed. Existing approaches mostly used pixel-level similarity~\cite{hinton2006reducing,inpainting}, but we argue that it is \textbf{not} an ideal objective because pursuing pixel-level similarity may neglect semantic information, but such information is important for a wide range of downstream tasks (\textit{e.g.}, object detection and instance segmentation). To facilitate richer semantic information to be learned, we advocate for \textbf{Se}mantic-\textbf{a}ware \textbf{Ge}neration (\textbf{SaGe}) that measures the generation score at the semantic level. This is done by introducing a self-supervised network that extracts features and thus measures the similarity between the original and generated images. As we would see in the experiments, equipping the evaluator with pre-learned semantics brings a major benefit and thus facilitates the encoder-decoder system to arrive at a better tradeoff between image recovery quality and semantic sensitivity. The comparison between SaGe and prior works is summarized in Table~\ref{tab:introduction}.

We evaluate SaGe-based visual representation learning (in short, SaGe) by pre-training the encoder-decoder system on ImageNet-1K and then transferring the encoder to a series of downstream tasks, including the linear classification test, nearest neighbor test, and semi-supervised test on ImageNet-1K, object detection and instance segmentation on MS-COCO, and semantic segmentation on Cityscapes. SaGe shows favorable performance compared to the existing approaches, and extensive ablation studies verify that the improvement indeed comes from the newly introduced proxy. More importantly, SaGe has the potential of replacing the contrastive/predictive learning, where the usage of intensive data augmentations can risk semantic inconsistency that harms the pre-trained models.

Overall, the contributions of this work are two-fold. First, we build a self-supervised learning framework that involves both discriminative and generative abilities, and validate its superior performance in a series of downstream tasks. Second, with extensive diagnostic and ablative experiments, we verify that the effectiveness of generation-based proxy depends on a semantic-aware evaluator, which we hope tp inspire future research in this routine.

\section{Related Work}
\label{sec:related}

Self-supervised visual representation learning is attracting increasing attentions, arguably because the increasing amount of unlabeled image data and the expensiveness of data annotation. The key to self-supervised learning is to design a pretext task, or equivalently a \textbf{proxy}, so that unlabeled images naturally satisfy. Below, we categorize prior work into three types of proxies.

The \textbf{geometry-based} proxies refer to the constraints based on the spatial relationship among image patches and/or the feature-level consistency across geometric transformations. Typical examples include~\cite{jigsaw} that required the network to solve a jigsaw puzzle so as to learn the spatial relationship between image patches~\cite{context,wei2019iterative}, \cite{rotation} that randomly rotated the image and asked the network to predict the angle of rotation, and ~\cite{colorization} that received a grayscale image and tried to recover the RGB version. Since geometries are sometimes predictable by low-level features, these approaches share a common weakness of learning effective high-level features, \textit{e.g.}, in the linear test on ImageNet, best performance usually appears upon mid-level features.

The \textbf{contrastive} proxy went one step further by generating different (most often, two) \textbf{views} of an image and assuming that the target network has the ability of extracting highly consistent features for these views. Intensive data augmentations~\cite{bachman2019learning,misra2020self} are often added to avoid the target network from learning naive parameters. There are mainly two paths to measure feature consistency. The first path is named \textbf{contrastive learning}~\cite{he2020momentum,wu2018unsupervised,zhuang2019local,oord2018representation,bachman2019learning,chen2020simple}, where view \#1 is put into a pool that contains a large number of distractors, and view \#2 is used as the query with the goal being to distinguishing view \#1 from all others -- this mechanism often appears as a classification task where each instance forms an independent class~\cite{dosovitskiy2015discriminative}, and the so-called memory bank~\cite{wu2018unsupervised,he2020momentum} significantly enlarged the pool size in a practical manner. There exist follow-up works that discussed various aspects of improving the learning performance, including encouraging locality-sensitive feature matching to avoid the conflict between data augmentation and image-level consistency~\cite{wang2021dense,li2021efficient,xie2021detco}, constructing prototypes~\cite{li2020prototypical,caron2020unsupervised} or inter-sample contrastive pairs~\cite{chen2020simple}, \textit{etc}. The second path is named \textbf{predictive learning}~\cite{grill2020bootstrap}, where the goal is to predict the feature extracted from view \#1 using that from view \#2. An important mechanism that avoids the model from collapsing is to use online and moving-averaged versions of the target network and perform the stop-gradient operator~\cite{grill2020bootstrap,chen2021exploring}. The predictive learning approaches enjoy elegance (\textit{e.g.}, the memory bank is no longer needed) but also suffer heavier computational overheads. Similarly, there are many efforts that tried to improve predictive learning from different aspects~\cite{tejankar2021isd,koohpayegani2021mean,zbontar2021barlow}.

\begin{figure*}
\begin{center}
\includegraphics[width=15.5cm]{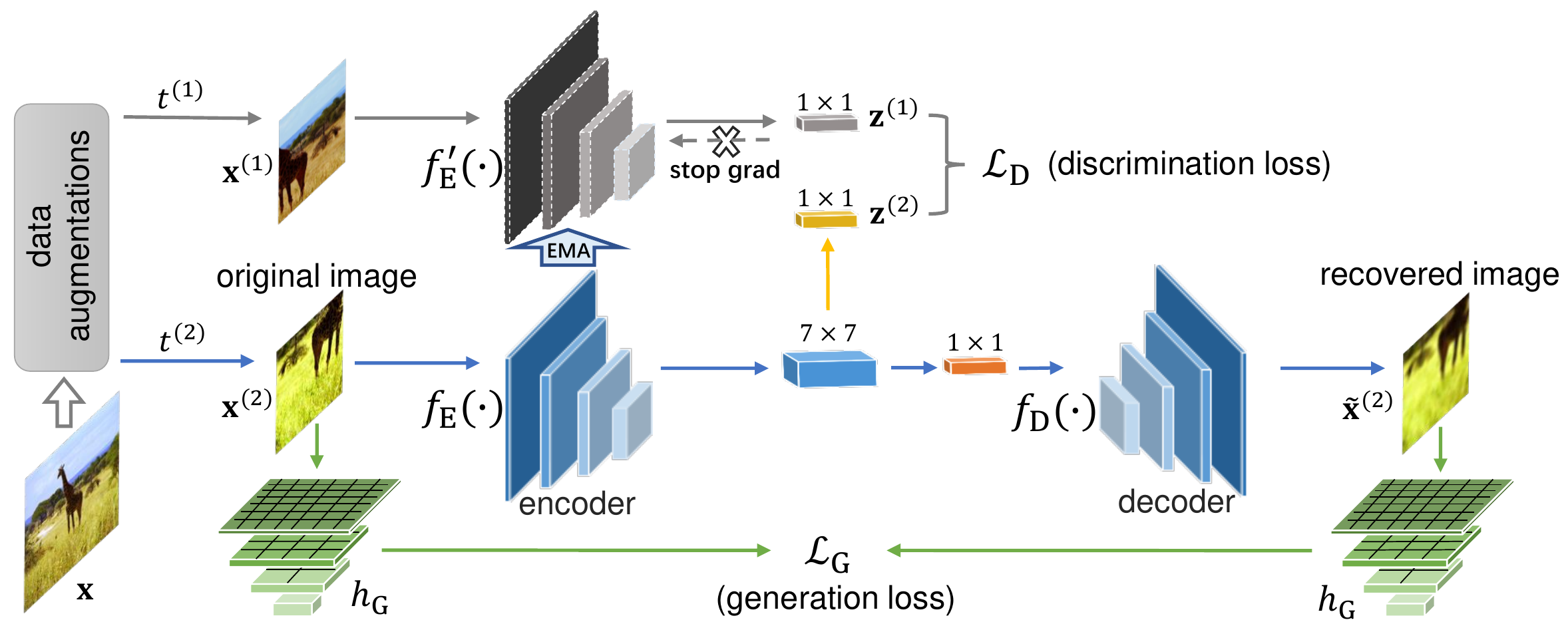}
\end{center}
\caption{The framework of SaGe. For brevity, we only show the loss terms related to the encoding-decoding procedure of $\mathbf{x}^{(2)}$, while there is a symmetric part that involves the encoding-decoding procedure of $\mathbf{x}^{(1)}$.}
\label{fig:framework}
\end{figure*}

The \textbf{generation} proxy assumes that the target network (which mainly encodes the input image into a compact vector) should have the ability of recovering the original image. The early efforts date back to research on generic data, where autoencoders~\cite{hinton2006reducing} where an auxiliary module named decoder is trained to recover the input data, and the variational version~\cite{kingma2013auto} that added constraints on the distribution of encoded data. The computer vision community later leveraged these ideas to learn visual representations~\cite{hinton2011transforming}, inheriting the encoder-decoder (\textit{a.k.a.}, discriminator-generator~\cite{donahue2016adversarial,dumoulin2016adversarially}) system. Examples include~\cite{vincent2008extracting,inpainting} that added noise to the input image and trained the network to recover it, \cite{zhang2017split} that reconstructed a part of input image from another part with the cross-channel features, \cite{aet} that minimized the input and output transformations in an end-to-end manner, \textit{etc}. Recently, generation-based learning shows inferior performance to contrastive or predictive learning -- in this paper, we reveal the reason behind this phenomenon to be the evaluation of generation quality neglecting semantics but mostly focusing on pixel-level similarity. As a side note, the generation proxy works better in the natural language processing community, \textit{i.e.}, the masked language modeling task~\cite{bert,lan2019albert,liu2019roberta}, arguably because the basic linguistic elements are clear and discrete, unlike the image pixels or patches being continuous yet may not reflect to complete semantics.

\section{Our Approach}
\label{sec:method}

In this section, we present the SaGe approach. We start with introducing the setting of self-supervised visual representation learning in Section~\ref{approach:setting}. Next, in Section~\ref{approach:framework}--Section~\ref{approach:design}, we elaborate the idea that considers both discrimination and generation, the semantic-aware generation, and design choices to wrap up the entire framework.

\subsection{Setting and Notations}
\label{approach:setting}

The goal of self-supervised visual representation learning is to learn a computational model $\mathbf{z}=f(\mathbf{x};\boldsymbol{\theta})$ from an unlabeled dataset $\mathcal{D}=\left\{\mathbf{x}_n\right\}_{n=1}^N$. In the deep learning era, $f(\mathbf{x};\boldsymbol{\theta})$ often appears as a deep neural network with $\boldsymbol{\theta}$ denoting the learnable parameters. Different from supervised settings, the output, $\mathbf{z}$, does not correspond to some particular semantics (\textit{e.g.}, the class label of $\mathbf{x}$), but serves as a compact representation (\textit{a.k.a.}, features) of $\mathbf{x}$. Hence, the downstream tasks can inherit these features for an efficient fine-tuning procedure to perform visual recognition.

Since no annotations are provided, most self-supervised visual representation learning involves defining a \textbf{proxy} (\textit{i.e.}, an annotation-free condition that all images shall satisfy) $\mathfrak{P}$ and transforming it into a loss function $\mathcal{L}(\cdot)$ to optimize $f(\mathbf{x};\boldsymbol{\theta})$. An example comes from the learning-by-predicting-rotation approach~\cite{rotation} where the proxy, denoted by $\mathfrak{P}^\mathrm{rot}$, demonstrates that the orientation of an image is always predictable regardless how it was rotated (\textit{e.g.}, clockwise by $90^\circ$). To use $\mathfrak{P}^\mathrm{rot}$ in self-supervised learning, one can add a random angle of rotation $r$ to any sample $\mathbf{x}_n$ from $\mathcal{D}$ and train the network to predict $r$, namely, $r=h(\mathbf{z}_n;\boldsymbol{\tau})=h(f(\mathbf{x};\boldsymbol{\theta});\boldsymbol{\tau})$, where $h(\mathbf{z}_n;\boldsymbol{\tau})$ is an auxiliary head that is used during the self-supervised learning procedure but discarded thereafter.

\subsection{A Discrimination-Generation Framework}
\label{approach:framework}

We construct a self-supervised visual representation learning framework that considers the abilities of both discrimination and generation, so that we can further diagnose and compare these two kinds of proxies. Here, by the \textbf{discrimination} proxy, $\mathfrak{P}^\mathrm{D}$, we refer to the ability of extracting consistent features for different views of an image; correspondingly, by the \textbf{generation} proxy, $\mathfrak{P}^\mathrm{G}$, we require the network to recover the original image from the compact features. The overall framework is illustrated in Figure~\ref{fig:framework}.

Mathematically, let $\mathbf{x}$ be an input image sampled from the unlabeled dataset, $\mathcal{D}$. Without loss of generality, we apply independent data augmentations to extract two views of $\mathbf{x}$, namely, $\mathbf{x}^{(1)}=t^{(1)}(\mathbf{x})$ and $\mathbf{x}^{(2)}=t^{(2)}(\mathbf{x})$, where $t^{(1)}(\cdot)$ and $t^{(2)}(\cdot)$ are sampled from a pre-defined set of transformation functions, $\mathcal{T}$. Data augmentation is crucial for constructing sufficiently different views of $\mathbf{x}^{(1)}$ and $\mathbf{x}^{(2)}$, otherwise extracting consistent features for them becomes a naive task. Both $\mathbf{x}^{(1)}$ and $\mathbf{x}^{(2)}$ are fed to an \textbf{encoder} (\textit{i.e.}, the target network) to obtain corresponding mid-level features, denoted by $\mathbf{z}^{(1)}=f(\mathbf{x}^{(1)};\boldsymbol{\theta}_\mathrm{E})$ and $\mathbf{z}^{(2)}=f(\mathbf{x}^{(2)};\boldsymbol{\theta}_\mathrm{E})$, where the subscript $\mathrm{E}$ stands for the `encoder'. $\mathbf{z}^{(1)}$ and $\mathbf{z}^{(2)}$ lay the foundation for discrimination, but to facilitate generation, we further feed $\mathbf{z}^{(1)}$ and $\mathbf{z}^{(2)}$ into a \textbf{decoder} to recover the input images. This is denoted by $\tilde{\mathbf{x}}^{(1)}={g(\mathbf{z}^{(1)};\boldsymbol{\theta}_\mathrm{D})}$ and $\tilde{\mathbf{x}}^{(2)}={g(\mathbf{z}^{(2)};\boldsymbol{\theta}_\mathrm{D})}$, where the subscript $\mathrm{D}$ stands for the `decoder'. Note that the decoder together with the parameters are not used in the downstream tasks, so we can refer to it as an auxiliary module that assists optimizing the encoder.

Based on the basic elements, namely, $\mathbf{x}^{(1)}$, $\mathbf{x}^{(2)}$, $\mathbf{z}^{(1)}$, $\mathbf{z}^{(2)}$, $\tilde{\mathbf{x}}^{(1)}$, $\tilde{\mathbf{x}}^{(2)}$, the key is to define two \textbf{proxies} for evaluating the abilities of discrimination and generation, respectively. We denote the loss functions for discrimination and generation as $\mathcal{L}_\mathrm{D}(\mathbf{z}^{(1)},\mathbf{z}^{(2)})$ and $\mathcal{L}_\mathrm{G}(\mathbf{x}^{(1)},\mathbf{x}^{(2)},\tilde{\mathbf{x}}^{(1)},\tilde{\mathbf{x}}^{(2)})$, respectively. Recent research offers efficient examples for $\mathcal{L}_\mathrm{D}(\mathbf{z}^{(1)},\mathbf{z}^{(2)})$ include putting $\mathbf{z}^{(2)}$ into an instance pool $\mathcal{B}$ (each instance forms a class) and compute the classification loss~\cite{chen2020simple,he2020momentum}, or using an auxiliary module $h_\mathrm{D}(\cdot)$ that transforms $\mathbf{z}^{(1)}$ to $h_\mathrm{D}(\mathbf{z}^{(1)})$ and measuring the distance between $h_\mathrm{D}(\mathbf{z}^{(1)})$ and $\mathbf{z}^{(2)}$ attaching with stop gradient mechanism~\cite{grill2020bootstrap, chen2020improved}. In comparison, existing examples for $\mathcal{L}_\mathrm{G}(\mathbf{x}^{(1)},\mathbf{x}^{(2)},\tilde{\mathbf{x}}^{(1)},\tilde{\mathbf{x}}^{(2)})$ mostly involves accumulating pixel-level inconsistency, \textit{e.g.},
\begin{equation}
\label{eqn:pixel-level}
\mathcal{L}_\mathrm{G}(\mathbf{x}^{(1)},\mathbf{x}^{(2)},\tilde{\mathbf{x}}^{(1)},\tilde{\mathbf{x}}^{(2)})=\|\mathbf{x}^{(1)}-\tilde{\mathbf{x}}^{(1)}\|_2+\|\mathbf{x}^{(2)}-\tilde{\mathbf{x}}^{(2)}\|_2,
\end{equation}
where the $\ell_2$-norm can be replaced by other (\textit{e.g.}, $\ell_p$-norm) metrics.

\begin{figure}[!t]
\centering
\includegraphics[width=7.5cm]{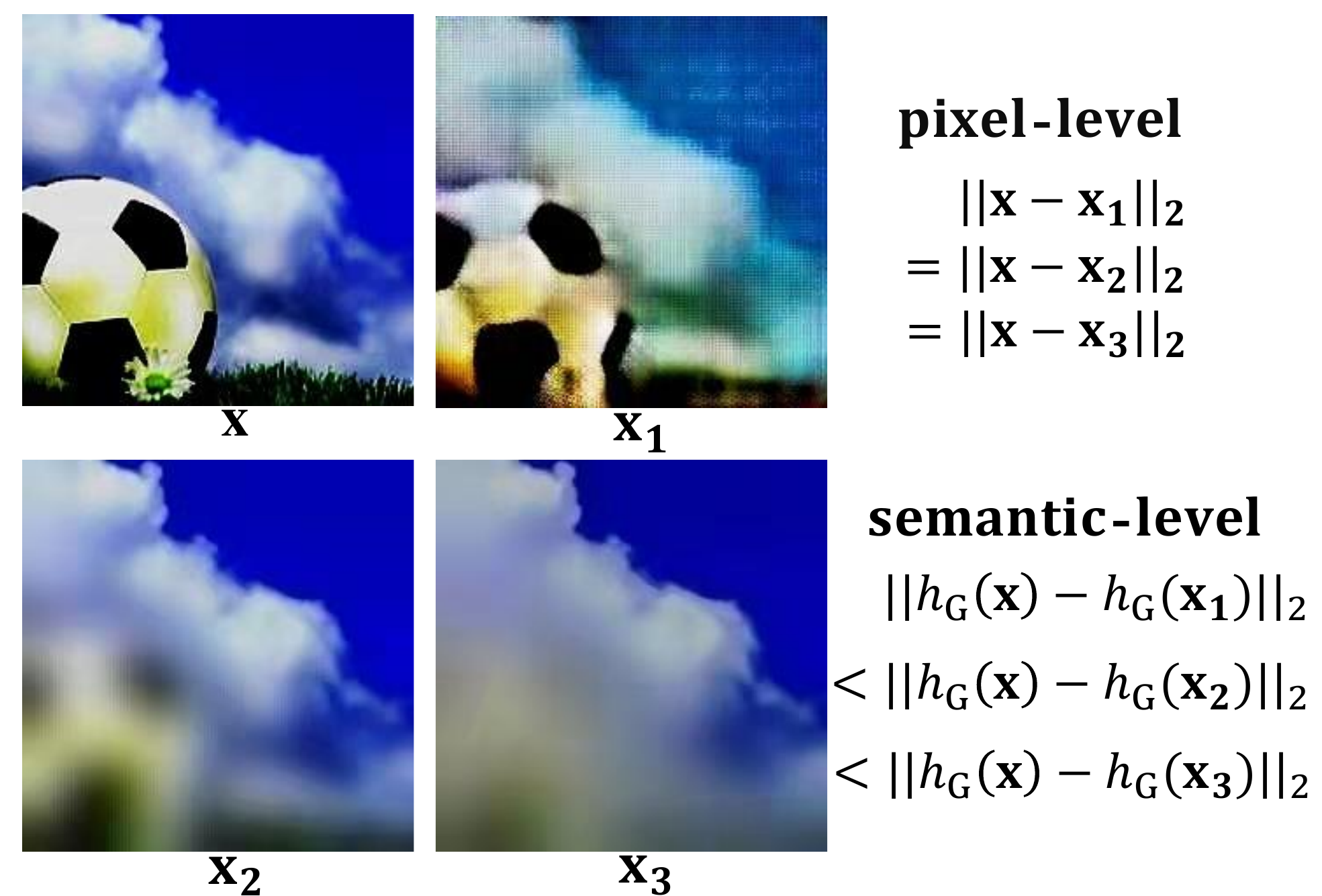}
\caption{An example showing that three candidates which are comparable in terms of pixel-level similarity but preserve significantly different extents of semantics. The pixel-level distance is measured by MSE, while the semantic-level distance is MSE calculated upon features extracted from the penultimate layer ($2\rm{,}048$D) of a pre-trained model by BYOL~\cite{grill2020bootstrap}.}
\label{fig:pixel-level}
\end{figure}

However, we point out that pixel-level inconsistency has the drawback of neglecting semantics and hence it is not a perfect objective to optimize. Figure~\ref{fig:pixel-level} shows an example, where the original image $\mathbf{x}$ is encoded into $\mathbf{z}$ and then decoded into three candidates that are equivalent in terms of $\ell_2$-norm distance, but one of them preserves almost complete semantics of the original image but other two does not. Producing such unsatisfying results implies that the encoder-decoder system fails to efficiently encode semantics into $\mathbf{z}$\footnote{It is also possible that the decoder does not work well to recover $\mathbf{x}$ from $\mathbf{z}$, but provided that in our design, the decoder is relatively lightweight (see Section~\ref{exp:details}), we mainly locate the issue in the encoder.}. Note that the goal is to transfer the encoder into downstream visual recognition tasks, it would be better to force it to preserve semantic information.

\subsection{Towards Semantic-Aware Generation}
\label{approach:generation}

The above analysis motivates us to introduce semantics into the generation proxy, or more specifically, into the loss function of $\mathcal{L}_\mathrm{G}(\mathbf{x}^{(1)},\mathbf{x}^{(2)},\tilde{\mathbf{x}}^{(1)},\tilde{\mathbf{x}}^{(2)})$. This is done by using another auxiliary module, $h_\mathrm{G}(\cdot)$, that extracts semantic-aware features from both the original and generated images. We name $h_\mathrm{G}(\cdot)$ as a semantic-aware evaluator. Equipped with $h_\mathrm{G}(\cdot)$, Eqn~\eqref{eqn:pixel-level} becomes
\begin{eqnarray}
\nonumber
\mathcal{L}_\mathrm{G}(\mathbf{x}^{(1)},\mathbf{x}^{(2)},\tilde{\mathbf{x}}^{(1)},\tilde{\mathbf{x}}^{(2)})=\\
\label{eqn:semantic-aware}
\|h_\mathrm{G}(\mathbf{x}^{(1)})-h_\mathrm{G}(\tilde{\mathbf{x}}^{(1)})\|_2+\|h_\mathrm{G}(\mathbf{x}^{(2)})-h_\mathrm{G}(\tilde{\mathbf{x}}^{(2)})\|_2.
\end{eqnarray}
Equivalently, the aim is to measure the distance between $\mathbf{x}^{(1)}$ and $\tilde{\mathbf{x}}^{(1)}$ after projecting them to the semantic space. The projector, $h_\mathrm{G}(\cdot)$, is the key component -- when $h_\mathrm{G}(\cdot)$ is an identity mapping function, Eqn~\eqref{eqn:semantic-aware} degenerates to Eqn~\eqref{eqn:pixel-level}. In this paper, we instantiate $h_\mathrm{G}(\cdot)$ as a pre-trained model \textbf{without} using labels (\textit{e.g.}, the model obtained from the BYOL algorithm~\cite{grill2020bootstrap} on ImageNet-1K). As shown in Figure~\ref{fig:pixel-level}, the candidates that are not distinguishable at the pixel level vary significantly at the semantic level, and the one that preserves most semantics has the lowest loss.

A discrimination-generation framework that contains $h_\mathrm{G}(\cdot)$ for semantic-aware evaluation is referred to as \textbf{S}emantic-\textbf{a}ware \textbf{Ge}neration (\textbf{SaGe}). Before entering design principles and experiments, we understand SaGe from the complementariness of its two branches, or equivalently, the discriminative and generative loss functions, $\mathcal{L}_\mathrm{D}(\mathbf{z}^{(1)},\mathbf{z}^{(2)})$ and $\mathcal{L}_\mathrm{G}(\mathbf{x}^{(1)},\mathbf{x}^{(2)},\tilde{\mathbf{x}}^{(1)},\tilde{\mathbf{x}}^{(2)})$. $\mathcal{L}_\mathrm{D}(\mathbf{z}^{(1)},\mathbf{z}^{(2)})$ pulls the features from two views together, aiming to eliminate the effect of $t^{(1)}(\cdot)$ and $t^{(2)}(\cdot)$. This proxy may cause the target network insensitive to image transformations. Although this strategy, in average, brings the benefit of learning image-level semantics, it can also cause the conflict between intensive data augmentation and image-level consistency, especially for complicated images~\cite{wang2021dense, pixpro}. We point out that the conflict is essentially the loss of view-specific information, and SaGe encourages $\mathbf{z}^{(1)}$ and $\mathbf{z}^{(2)}$ to preserve the view-specific information (so that the decoder can recover the original image), hence alleviating the conflict and improving the learning quality.

\subsection{Design Choices and Discussions}
\label{approach:design}

We instantiate SaGe upon the baseline of BYOL~\cite{grill2020bootstrap}, yet this does not obstacle its application to other methods, \textit{e.g.}, MoCo~\cite{he2020momentum}. The overall loss function is written as:
\begin{equation}
\mathcal{L}_\mathrm{overall}=\mathcal{L}_\mathrm{D}(\mathbf{z}^{(1)},\mathbf{z}^{(2)})+\lambda\cdot\mathcal{L}_\mathrm{G}(\mathbf{x}^{(1)},\mathbf{x}^{(2)},\tilde{\mathbf{x}}^{(1)},\tilde{\mathbf{x}}^{(2)}),
\end{equation}
where $\lambda$ is the balancing coefficient. Since $h_\mathrm{G}(\cdot)$ is a hierarchical function, we compute the distance between $\mathbf{x}^{(1)}$ and $\tilde{\mathbf{x}}^{(1)}$, and similarly, between $\mathbf{x}^{(2)}$ and $\tilde{\mathbf{x}}^{(2)}$, on multiple levels, including the pixel level. We borrow $h_\mathrm{G}(\cdot)$ from the pre-trained model of BYOL that runs on ImageNet for $300$ epochs, and freeze its parameters during the SaGe training procedure. After the pre-training, all of $g(\cdot)$, $h_\mathrm{D}(\cdot)$ and $h_\mathrm{G}(\cdot)$ are discarded and $f(\cdot)$ is preserved and transferred to downstream tasks.

Below, we briefly discuss some design choices.

\vspace{0.1cm}\noindent
$\bullet$\hspace{0.2cm}\textbf{Why not minimizing $\|h_\mathrm{G}(\mathbf{x}^{(1)})-h_\mathrm{G}(\tilde{\mathbf{x}}^{(2)})\|_2$ or $\|h_\mathrm{G}(\mathbf{x}^{(2)})-h_\mathrm{G}(\tilde{\mathbf{x}}^{(1)})\|_2$?} This is mainly due to the potential difference between $\mathbf{x}^{(1)}$ and $\mathbf{x}^{(2)}$, caused by intensive data augmentations. A straightforward example lies in random crop, where $\mathbf{x}^{(1)}$ and $\mathbf{x}^{(2)}$ may correspond to different regions of $\mathbf{x}$, and thus the above task of minimization is not reasonable. In addition, as elaborated in the previous part, we hope the encoder to preserve view-specific information, which aligns with intra-view learning, \textit{i.e.}, minimizing $\|h_\mathrm{G}(\mathbf{x}^{(1)})-h_\mathrm{G}(\tilde{\mathbf{x}}^{(1)})\|_2$ and $\|h_\mathrm{G}(\mathbf{x}^{(2)})-h_\mathrm{G}(\tilde{\mathbf{x}}^{(2)})\|_2$.

\vspace{0.1cm}\noindent
$\bullet$\hspace{0.2cm}\textbf{Why freezing $h_\mathrm{G}(\cdot)$ during the SaGe training procedure?} This is mainly to avoid degeneration, \textit{i.e.}, $h_\mathrm{G}()$ tending to extract very similar features for any $\mathbf{x}$ and hence both $\|h_\mathrm{G}(\mathbf{x}^{(1)})-h_\mathrm{G}(\tilde{\mathbf{x}}^{(1)})\|_2$ and $\|h_\mathrm{G}(\mathbf{x}^{(2)})-h_\mathrm{G}(\tilde{\mathbf{x}}^{(2)})\|_2$ become small regardless of the \textbf{real} similarity between the original and recovered images. In practice, switching off this option brings consistent accuracy drop ($>20\%$) on the ImageNet-1K linear classification test.

\vspace{0.1cm}\noindent
$\bullet$\hspace{0.2cm}\textbf{Why recovering the original image is required?} An equivalent version: why not minimizing $\|h_\mathrm{G}(\mathbf{x}^{(\cdot)})-\mathbf{z}^{(\cdot)}\|_2$, but minimizing $\|h_\mathrm{G}(\mathbf{x}^{(\cdot)})-h_\mathrm{G}(g(\mathbf{z}^{(\cdot)}))\|_2$? First, note that $\mathbf{z}^{(\cdot)}=f(\mathbf{x}^{(\cdot)})$ and,
based on the previous question, $h_\mathrm{G}(\cdot)$ is frozen. So, minimizing $\|h_\mathrm{G}(\mathbf{x}^{(\cdot)})-\mathbf{z}^{(\cdot)}\|_2$ implies forcing $f(\cdot)$ to mimic the behavior of $h_\mathrm{G}(\cdot)$, which limits the potential of $f(\cdot)$. 

\section{Experiments}
\label{sec:experiment}

\subsection{Experimental Settings}
\label{exp:details}

We evaluate SaGe by pre-training it on the unlabeled ImageNet-1K~\cite{russakovsky2015imagenet} training set and transferring the encoder to other downstream visual recognition tasks. This protocol is widely used in the community. The downstream datasets and tasks include linear classification test, K-nearest-neighbor test, and semi-supervised test on ImageNet-1K, object detection and instance segmentation on MS-COCO~\cite{lin2014coco}, instance segmentation and semantic segmentation on Cityscapes~\cite{cordts2016cityscapes}.

We employ a $50$-layer ResNet~\cite{he2016deep} as the encoder. Standard data augmentations as in~\cite{grill2020bootstrap} are used to generate $\mathbf{x}^{(1)}$ and $\mathbf{x}^{(2)}$ with $224\times224$ pixels (in RGB, \textit{i.e.}, having $3$ channels). The encoder output $\mathbf{z}^{(\cdot)}$ with a dimensionality of $7\times7\times2048$. The feature is sent to two branches for discrimination and generation proxies, respectively. For the discrimination branch, we borrow the setting from BYOL~\cite{grill2020bootstrap}, which adds a $2$-layer MLP on top of the average-pooled feature ($2\rm{,}048$D) and obtains a $256$D embedding vector. For the generative branch, we first apply a $7\times7$ convolution to reduce the dimensionality of $\mathbf{z}^{(\cdot)}$ to $1\times1\times512$, and then decode it using a series of deconvolution layers with a up-sampling rate of $2$ (the first with rate of $7$) and the output dimensionality being $7\times7\times512$, $14\times14\times256$, $28\times28\times128$, $56\times56\times64$, $112\times112\times32$, and $224\times224\times16$, respectively. The final one is further propagated through a $1\times1$ convolution to output $\tilde{\mathbf{x}}^{(\cdot)}$ with a dimensionality of $224\times224\times3$.

The optimization mainly follows the convention. A moving-average copy of the encoder is updated with a momentum which increases from $0.9$ to $1.0$ during the training procedure. To avoid collapse, the stop-gradient operation is used so that the copy is not updated by gradients. The encoder is optimized using LARS~\cite{huo2021large} optimizer with a momentum of $0.9$, a learning rate of $4.8$, and a weight decay of $10^{-6}$. The decoder is trained using Adam~\cite{adam} with a learning rate of $0.048$. The encoder's learning rate is gradually decayed to $0$ in a cosine annealing schedule while the decoder's learning rate remains unchanged. Unless otherwise specified, the evaluator is another ResNet-50 pre-trained using BYOL on ImageNet-1K for $300$ epochs, which get a linear evaluation result of $73.1\%$ ($72.5$ is reported in the original paper). All the experiments are conducted on 32 NVIDIA Tesla-V100 GPUs.
It takes about 4 days to train a 300 epochs model.

\begin{table}[!t]
\fontsize{9.5}{11.0}\selectfont
\centering
\setlength{\tabcolsep}{2.5mm}
\begin{tabular}{l|c|cc|c}
\toprule
Method   & epochs  & Top-1  & Top-5 & $K$-NN \\
\midrule
PIRL~\cite{misra2020self} &800&63.6 &--&--\\
CPC-v2~\cite{henaff2019data}&200&63.8  &85.3&--\\
PCL~\cite{li2020prototypical}&200 &61.5 &--&--\\
SimCLR-v2~\cite{chen2020big} &800&71.7  &89.0&60.7\\
MoCo-v2~\cite{chen2020improved} &800 &71.1 &--&61.9\\
BYOL~\cite{grill2020bootstrap} &300 &72.7  &91.6&--\\
BYOL~\cite{grill2020bootstrap} &1000 &74.3  &91.6&64.8\\
\midrule
\textbf{SaGe} (short)  &300  &74.5  &91.8  &68.2\\
\textbf{SaGe} (long)  &800  &75.0  &92.1 & 68.7\\
\midrule
SwAV$^\dagger$~\cite{caron2020unsupervised} &300 &74.1 &-- &65.4\\
SwAV$^\dagger$~\cite{caron2020unsupervised} &800 &75.3 &-- &65.7\\
DINO$^\dagger$~\cite{caron2021emerging}  &300  &74.5 &-- &65.6\\
DINO$^\dagger$~\cite{caron2021emerging}  &800 &75.3 &-- &67.5\\
\bottomrule
\end{tabular}
\caption{Classification accuracy (\%) under the linear evaluation and $K$-nearest-neighbor test on ImageNet-1K. All the candidates are based on the ResNet-50 backbone. The entries marked with $\dagger$ indicate that multi-crop augmentation has been used, which typically improves top-1 accuracy by $1\%$--$2\%$. We report the 20-NN scores in the last column following the convention.}
\label{tab:lincls_imagenet1K}
\end{table}

\subsection{Performance on Downstream Tasks}

\noindent
$\bullet$\hspace{0.2cm}\textbf{Linear evaluation on ImageNet-1K.} Following the standard setting, we inherit the pre-trained encoder, $f(\cdot)$, average-pool the output features into a $2\rm{,}048$-dimensional vector, and build two fully-connected layers upon it. We only fine-tune the FC layers on the entire training set of ImageNet-1K with the standard settings. As the results summarized in Table~\ref{tab:lincls_imagenet1K}, SaGe with either $300$ or $800$ pre-training epochs outperforms all competitors without multi-crop augmentation, and is on par with the results upon multi-crop augmentation. In particular, with $300$ pre-training epochs, SaGe outperforms the BYOL baseline with $800$ pre-training epochs, implying its efficiency in visual representation learning.

\vspace{0.1cm}\noindent
$\bullet$\hspace{0.2cm}\textbf{$K$-nearest-neighbor test on ImageNet-1K.} Still, we make use of the pre-trained $f(\cdot)$ to extract a $2\rm{,}048$-dimensional vector, and then directly use it to retrieve $K$ training samples with the smallest distances. We report the 20-NN scores and the results are shown in the last column of Table~\ref{tab:lincls_imagenet1K}. In this scenario that fine-tuning is absent, the advantage of SaGe becomes more significant -- it even outperforms the competitors with multi-crop augmentation, validating the effectiveness of view-specific information.

\begin{table}[!t]
\fontsize{9.5}{11.0}\selectfont
\centering
\setlength{\tabcolsep}{2.0mm}
\begin{tabular}{l|cc|cc}
\toprule
\multirow{2}{*}{Method} & \multicolumn{2}{c|}{$1\%$ labels} & \multicolumn{2}{c}{$10\%$ labels} \\
\multicolumn{1}{c|}{}            & Top-1  & Top-5    & Top-1  & Top-5 \\\midrule
Supervised                      & 25.4   & 48.4     & 56.4   & 80.4  \\ \midrule
PIRL~\cite{misra2020self}        & 30.7   & 57.2     & 60.4   & 83.8  \\
PCL~\cite{li2020prototypical}    & --     & 75.6     & --     & 86.2 \\
SimCLR~\cite{chen2020simple}     & 48.3   & 75.5     & 65.6   & 87.8 \\
MoCo-v2~\cite{chen2020improved}  & 52.4   & 78.4     & 65.3   & 86.6 \\
SwAV\cite{caron2020unsupervised} & 53.9   & 78.5     & 70.2   & 89.9  \\
\midrule
\textbf{SaGe} (800 epochs)  &57.8   &81.5     &68.7   &88.5   \\
\midrule
FixMatch$^{\dagger}$~\cite{sohn2020fixmatch} & --     & --       & 71.5   & 89.1  \\
SimCLR-v2$^\ddagger$\cite{chen2020big}      & 57.9   & 82.5     & 68.4   & 89.2 \\
\bottomrule
\end{tabular}
\caption{Classification accuracy (\%) under the semi-supervised learning protocol on ImageNet-1K. The entry with $\dagger$ indicates that RandAugment has been used and $\ddagger$ indicates the method is specially designed for semi-supervised task.}
\label{tab:semi_imagenet}
\end{table}

\vspace{0.1cm}\noindent
$\bullet$\hspace{0.2cm}\textbf{Semi-supervised evaluation on ImageNet-1K.} The architecture is same as linear evaluation, but only $1\%$ or $10\%$ of training set are labeled and the entire network including the pre-trained encoder can be fine-tuned. Results in Table~\ref{tab:semi_imagenet} show the competitive performance of SaGe -- among all approaches with standard data augmentations (random cropping and horizontal flipping), it is the best and second-best on $1\%$ and $10\%$ labels, respectively, yet its accuracy is on par with the approaches applying RandAugment~\cite{RandAugmentation} or specially designed for semi-supervised task.

\begin{table}[!t]
\centering
\fontsize{9.5}{11.0}\selectfont
\setlength{\tabcolsep}{2.0mm}
\begin{tabular}{l|cc|c}
\toprule
\multirow{2}{*}{Method} &\multicolumn{2}{c|}{instance seg.} & semantic seg. \\ 
                         & $\mathrm{AP}^{\mathrm{bb}}$ &$\mathrm{AP}_{50}^{\mathrm{bb}}$
                         & mIoU\\ \midrule
Supervised &       32.9     &      59.6  & 74.6    \\ \midrule
MoCo-v1~\cite{he2020momentum} &   32.3    &    59.3  & 75.3               \\
MoCo-v2~\cite{chen2020improved} &   33.1   &     60.1     & 75.1             \\
HCL~\cite{HCL} & 33.6          &   60.8  &  75.5                \\ 
DenseCL~\cite{wang2021dense} & --          &  --  &  75.7                \\ \midrule
\textbf{SaGe} (800 epochs)  &   33.9   &     61.3    & 76.9            \\ 
\bottomrule
\end{tabular}
\caption{Segmentation accuracy (\%) on the Cityscapes dataset. All the candidates are based on the ResNet-50 backbone. Following MoCo, the instance segmentation is built upon Mask R-CNN with FPN and semantic segmentation is based on FCN.}
\label{tab:cityscapes}
\end{table}

\vspace{0.1cm}\noindent
$\bullet$\hspace{0.2cm}\textbf{Object detection and instance segmentation on MS-COCO.} We then transfer the pre-trained models for object detection and instance segmentation on the MS-COCO dataset~\cite{lin2014coco}. Following~\cite{he2020momentum}, we use Mask R-CNN~\cite{he2017mask} with the FPN head~\cite{lin2017feature} and fine-tune both the backbone and head under the $1\times$ schedule. The train2017 and val2017 sets are used for training and testing, respectively. As shown in Table~\ref{tab:coco}, SaGe consistently outperforms MoCo-v2 and the BYOL baseline. Interestingly, SaGe outperforms DenseCL~\cite{wang2021dense} and self-EMD~\cite{liu2020self}, two self-supervised learning approaches that were particularly designed for object-level prediction, showing that the generation task is an alternative way of improving object-level description.

\vspace{0.1cm}\noindent
$\bullet$\hspace{0.2cm}\textbf{Semantic segmentation on Cityscapes.}
Lastly, we transfer the pre-trained models to the Cityscapes dataset. For instance segmentation, we use Mask R-CNN with the FPN head and perform a 2$\times$ training schedule; for semantic segmentation, we use the FCN head~\cite{long2015fully}. Table~\ref{tab:cityscapes} shows a similar trend as of MS-COCO experiments, showing the advantage of SaGe. This is easily interpreted. Discrimination-based learning suppresses view-specific information, but segmentation is sensitive to views. Generation-based learning complements the information loss and thus enhances the pre-trained models in such downstream tasks.

\begin{table*}[]
\renewcommand\arraystretch{1.0}
\setlength{\tabcolsep}{2.0mm}
\fontsize{9.0}{11}\selectfont
\centering
\begin{tabular}{l|cccccc|cccccc}
\toprule
\multirow{2}{*}{Method} & \multicolumn{6}{c|}{Mask R-CNN, R50-FPN, detection}  &\multicolumn{6}{c}{Mask R-CNN, R50-FPN, segmentation} \\
& AP$^{\mathrm{bb}}$  & AP$^{\mathrm{bb}}_{\mathrm{50}}$ & AP$^{\mathrm{bb}}_{\mathrm{75}}$ & AP$_{\mathrm{S}}$  & AP$_{\mathrm{M}}$ &AP$_\mathrm{L}$ &AP$^{\mathrm{mk}}$  &AP$^{\mathrm{mk}}_{\mathrm{50}}$  &AP$^{\mathrm{mk}}_{\mathrm{75}}$  &AP$_\mathrm{S}$  &AP$_\mathrm{M}$ &AP$_\mathrm{L}$ \\ \midrule

Supervised                  &38.9  &59.6  &42.0  &23.0  &42.9  &49.9    &35.4  &56.5  &38.1  &17.5  &38.2 &51.3     \\ \midrule
MoCo v2 ~\cite{chen2020improved} &39.2  &59.9 &42.7 &23.8 &42.7 &50.0    &35.7  &56.8  &38.1  &17.8  &38.1  &50.5   \\
BYOL ~\cite{grill2020bootstrap} &39.9 &60.2 &43.2 &23.3 &43.2 &52.8   &-- &--	&-- &-- &-- &--\\
HCL~\cite{HCL}  &40.0  & 60.6  &43.8  & --  & -- & -- &36.4  &57.6  & 39.1 &--  &--  &--\\
DenseCL~\cite{wang2021dense}  &40.3 &59.9 &44.3 &-- &-- &--    &36.4 &57.0 &39.2 &-- &-- &-- \\ 
self-EMD~\cite{liu2020self}  &40.0   &60.4  &44.0  &23.5  &43.8  &52.2   &-- &--	&-- &-- &-- &--\\
ORL~\cite{ORL} &40.3  &60.2  &44.4  &-- &-- &--  &36.3  &57.3  & 38.9  &-- &-- &-- \\ \midrule

\textbf{SaGe} (300 epochs) &40.2  &61.7  &43.8  &23.8  &44.1   &51.9  &36.7  &58.3  &39.2  &17.6  &39.5  &52.5 \\
\textbf{SaGe} (800 epochs) &40.8  &62.4  &44.8   &25.1  &44.5  &52.5   &37.2  &59.0  &40.1  &18.5  &40.0  &53.2  \\ 
\bottomrule
\end{tabular}
\caption{Object detection and instance segmentation APs (\%) on the MS-COCO dataset.}
\label{tab:coco}
\end{table*}

\begin{table}[]
\setlength{\tabcolsep}{2.0mm}
\fontsize{10}{11}\selectfont
\centering
\begin{tabular}{x{16}|x{16}|x{46}x{42}|x{24}}
\toprule
\multirow{2}{*}{ID} & \multirow{2}{*}{$\mathcal{L}_\mathrm{D}$} & \multicolumn{2}{c|}{$\mathcal{L}_\mathrm{G}$} & \multirow{2}{*}{Top-1}\\ \cline{3-4}
 & & $h_\mathrm{G}(\cdot) $ & epochs \\ \midrule
1 & \xmark & \xmark & -- & 0.9 \\
2 & \xmark & ResNet-18  & \textit{random}  &21.7 \\
3 & \xmark &ResNet-50       &300     &70.3 \\ 
4 & \cmark & \xmark & --  & 66.8 \\
5 & \cmark &ResNet-18       & \textit{random}       & 67.3 \\
6 & \cmark &ResNet-18       &1       & 67.7 \\
7 & \cmark &ResNet-18        &100      & 68.9  \\
8 & \cmark &ResNet-18        &300     & 70.1  \\
9 & \cmark &ResNet-50       &300     &71.9 \\
\bottomrule
\end{tabular}
\caption{Classification accuracy (\%) of linear evaluation under different configurations, where \#4 is the BYOL baseline ($100$ epochs), and \#9 is the complete version of SaGe.}
\label{tab:knowledge}
\end{table}

\subsection{The Sources of Knowledge}

This subsection answers an essential question: \textbf{what are key proxies that support the learning procedure of SaGe?} For this purpose, we provide an ablation study that involves different combinations of discrimination and generation proxies in Table~\ref{tab:knowledge}. Throughout the remaining part of this paper, we use a shorter schedule that pre-trains SaGe for $100$ epochs on ImageNet-1K.

On the one hand, the effect of $\mathcal{L}_\mathrm{D}$, \textbf{the discrimination proxy}, reflects in the direct comparisons between (\#1 vs. \#4), (\#2 vs. \#5), and (\#3 vs. \#9) of Table~\ref{tab:knowledge}. When there is no $\mathcal{L}_\mathrm{G}$, $\mathcal{L}_\mathrm{G}$ is very weak, and $\mathcal{L}_\mathrm{G}$ is strong, introducing $\mathcal{L}_\mathrm{D}$ brings $65.9\%$, $45.6\%$, and $1.6\%$ accuracy gains, respectively, showing a marginal effect.

On the other hand, the effect of $\mathcal{L}_\mathrm{G}$, \textbf{the generation proxy}, is revealed by the direct comparisons between (\#1--\#3) and (\#4--\#9) of Table~\ref{tab:knowledge}. When $\mathcal{L}_\mathrm{D}$ is absent, using $\mathcal{L}_\mathrm{G}$ itself can force the network to learn semantics. On the other hand, $\mathcal{L}_\mathrm{G}$ still contributes even when a strong $\mathcal{L}_\mathrm{D}$ (\textit{i.e.}, BYOL) is present. Interestingly, when we compare \#4 against \#5 and \#6, a random evaluator and an evaluator pre-trained for merely $1$ epoch bring accuracy gains of $0.5\%$ and $0.9\%$, respectively, though larger gains are obtained with stronger evaluators.

Integrating the above comparisons, we learn the lesson that discrimination and generation are indeed complementary in self-supervised visual representation learning -- the former is good at learning image-level semantics, and the latter complements it by enforcing view-specific recovery. SaGe goes one step further by amending the goal of generation -- instead of recovering every single pixel, it should focus on capturing the semantics of the original image. This is more reasonable, since the encoder often produces a low-dimensional feature vector (\textit{e.g.}, in SaGe, a $512$-D vector that is $294\times$ smaller than the input image) -- it is expected to lose details, but deemed acceptable if most of semantics have been preserved.

\begin{figure*}
\centering
\includegraphics[width=17.4cm]{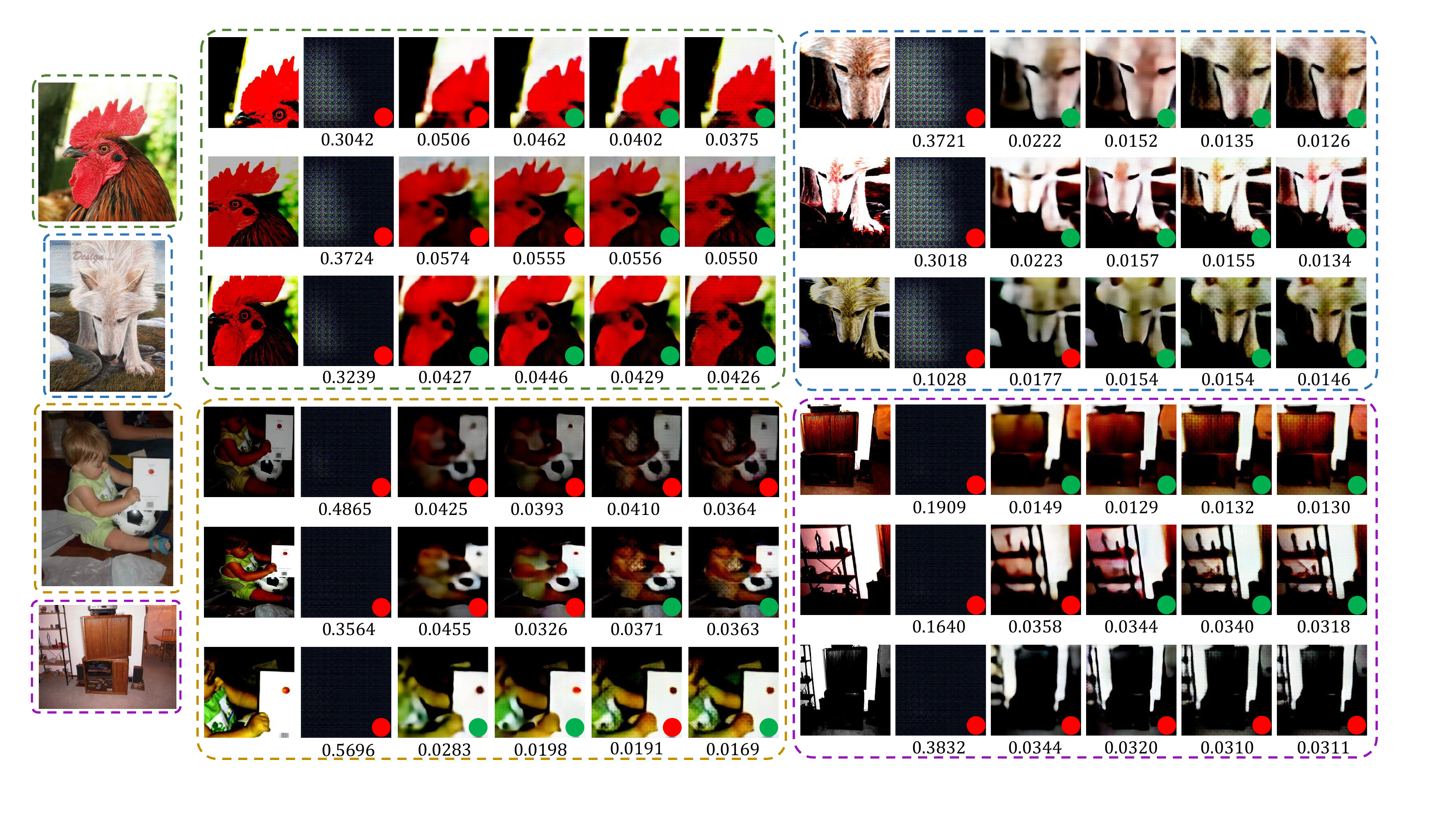}
\caption{Examples of generated images. The very first column shows four examples from ImageNet-1K, each of which corresponds to a block (bounded by the same color) in the right-hand side. In each block, the first column shows three views sampled from the corresponding image, and the remaining five columns correspond to the five options in Table~\ref{tab:generation}, namely, without MSE, MSE only, MSE with random R-18, MSE with pre-trained R-18, and MSE with pre-trained R-50, respectively. Green and red dots indicate that the corresponding case passes the perceptual test or not, and the MSE is offered below each case. \textit{This figure is best viewed by zooming into details.}}
\label{fig:visualization}
\end{figure*}

\begin{table}[]
\setlength{\tabcolsep}{1.8mm}
\fontsize{9.5}{11} \selectfont
\centering
\begin{tabular}{x{16}|x{20}|x{20}x{28}|x{24}|x{28}|x{24}}
\toprule
\multirow{3}{*}{ID} & \multicolumn{3}{c|}{$\mathcal{L}_\mathrm{G}$}  &\multirow{3}{*}{\stackbox[c][c]{Top-1\\linear\\acc.}}  &\multirow{3}{*}{\stackbox[c][c]{Top-5\\precept.\\acc.}} &\multirow{3}{*}{MSE} \\ \cline{2-4}
& \multirow{2}{*}{\stackbox[c][c]{MSE\\loss}}   & \multicolumn{2}{c|}{higher levels} & & & \\ \cline{3-4}
& & $h_\mathrm{G}(\cdot)$ & epochs & & & \\ \midrule
1 & \xmark &R-50  &300       &71.6   & 0.5 &0.3435 \\
2 & \cmark & \xmark & -- &67.2   & 40.5 &0.0297 \\
3 & \cmark &R-18  & \textit{random} &67.3   & 42.6 &0.0290 \\
4 & \cmark &R-18  &300       &70.1   & 50.4 &0.0287 \\
5 & \cmark &R-50  &300       &71.9   & 52.3 & 0.0285\\
\bottomrule
\end{tabular}
\caption{The accuracy (\%) of linear classification and perceptual test as well as the MSE for different models. All the entries are built upon the BYOL baseline, \textit{i.e.}, $\mathcal{L}_\mathrm{D}$ is present. BYOL, producing a $66.8\%$ linear classification accuracy, is not displayed since it does not generate images for the perceptual test.}
\label{tab:generation}
\end{table}

\subsection{Perceptual Test and Visualization}

In the last part, we investigate the quality of generated images. Besides using different options to encode and decode different views, as we show in Figure~\ref{fig:visualization}, we also perform a perceptual test. The setting is simple. We sample an image (and also, a view) and feed it through the encoder-decoder architecture. Then, we use a ResNet-50 model that was trained on ImageNet-1K \textbf{with labels}, and check if the recovered image is correctly classified\footnote{We use top-5 accuracy because the recovered images often suffer information loss in details, which causes accuracy drop especially for fine-grained recognition.}. This is an alternative way to test if the recovered image retains the original semantics. We name this task as the \textbf{perceptual test} -- note that it is not possible for purely discrimination-based learning approaches to perform this task.

Results are summarized in Table~\ref{tab:generation}. We first recall the readers that $h_\mathrm{G}(\cdot)$ contains both an MSE and higher-level loss terms -- the MSE loss, though with a small coefficient, is crucial for generating a visually meaningful image. From the comparison between \#1 and \#5 of Table~\ref{tab:generation}, we learn that an MSE-absent system is unable to pass the perceptual test with an accuracy barely above the random baseline, though the absence of MSE only causes a slight accuracy drop of $0.3\%$ in the linear test. In contrast, by comparing \#2 and \#5, we discover another phenomenon that missing the constraints of higher-level similarity results in a moderate drop in both tests. Interestingly, the MSE of \#2 is lower than that of \#5, indicating that pursuing pixel-level accuracy does not even necessarily benefit the perceptual test. The comparison between \#3--\#5 reveal an expected conclusion -- stronger evaluators produce stronger pre-trained models in both the linear and perceptual tests, while the MSE is mostly unaffected.

We hope the above studies deliver a new message to the community that the abilities of discrimination and generation are not bound. Although discrimination seems more important for a wide range of visual recognition tasks, we advocate for more attentions to generation which may benefit low-level vision tasks such as image denoising and/or high-quality image synthesis. We enhance the understanding using the examples shown in Figure~\ref{fig:visualization}.

\section{Conclusions}

In this paper, we present SaGe, a self-supervised visual representation learning approach which considers both discrimination and generation. Different from prior work, the generation branch is guided by a semantic-aware evaluator, which facilitates semantically meaningful information to be retained. Both quantitative and qualitative studies, including a perceptual test, validate the favorable ability of SaGe in transfer learning as well as preserving view-specific information.

\vspace{0.1cm}\noindent
\textbf{Limitations of this work.} SaGe suffers two issues. First, the requirement of a pre-trained evaluator increases the design and computational complexities. Second, the conflict between data augmentation and image-level consistency still exists, yet we are unready to remove the discrimination part. In the future, we will investigate the potential of SaGe in more challenging self-supervised learning scenarios, \textit{e.g.}, masked image modeling~\cite{bao2021beit} and/or reducing the dimensionality of internal representations, $\mathbf{z}$, towards a higher data compression ratio.

{\small
\bibliographystyle{ieee_fullname}
\bibliography{egbib}
}

\newpage
\appendix

\section{Details of the Evaluator}

\textbf{\textit{This part complements Sections~3.3, 3.4 and~4.1 in elaborating the implementation of the proposed evaluator.}}

Based on a pre-trained evaluator (\textit{e.g.}, ResNet-50 as used in most experiments), we extract multi-level features from both the original and recovered images and compute the $\ell_2$ loss between them. Specifically, four levels of features are extracted with the spatial resolutions being $28\times28$, $14\times14$, $7\times7$, and $1\times1$, and the numbers of channels being $512$, $1024$, $2048$, and $2048$, respectively. Note that the pixel-level $\ell_2$ distance is also computed as an additional term, which we refer to as the MSE loss. We denote these losses as $\mathcal{L}_\mathrm{MSE}$, $\mathcal{L}_{28\times28}$, $\mathcal{L}_{14\times14}$, $\mathcal{L}_{7\times7}$, and $\mathcal{L}_{1\times1}$, and hence the overall generative loss, $\mathcal{L}_\mathrm{G}$, is written as
\begin{eqnarray}
\nonumber
\mathcal{L}_\mathrm{G}=\lambda_\mathrm{MSE}\cdot\mathcal{L}_\mathrm{MSE}+\lambda_\mathrm{28\times28}\cdot\mathcal{L}_\mathrm{28\times28}+\lambda_\mathrm{14\times14}\cdot\mathcal{L}_\mathrm{14\times14}\\
+\lambda_\mathrm{7\times7}\cdot\mathcal{L}_\mathrm{7\times7}+\lambda_\mathrm{1\times1}\cdot\mathcal{L}_\mathrm{1\times1},
\end{eqnarray}
where all the balancing coefficients are set to be $0.1$.


\setcounter{figure}{3}
\begin{figure}
\centering
\includegraphics[width=6.5cm]{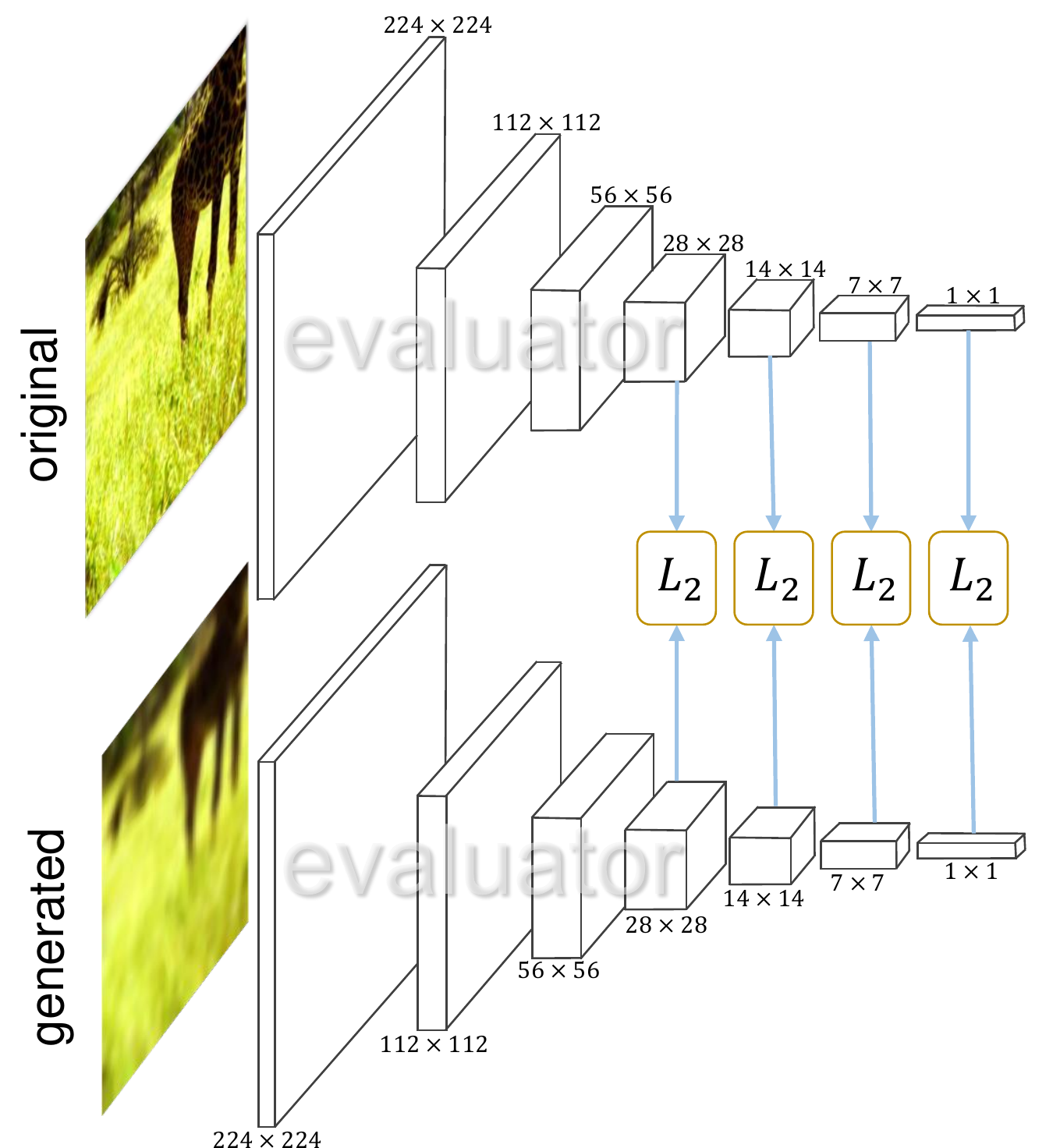}
\caption{The details of the architecture of $h_\mathrm{G}(\cdot)$. The goal is to reduce the distance of multi-level features between the original and recovered images.}
\label{fig:perceptual}
\end{figure}

\section{Details of Data Augmentation}

\textbf{\textit{This part complements the details described in Section~4.1 of the main article.}}

During the pre-training stage, we use the same data augmentations with ~\cite{grill2020bootstrap}, including random cropping, left-right flip, color jitting, Gaussian bluring, and solarization. The normalized mean and variance of the image are $[0.5,0.5,0.5]$ and $[0.5,0.5,0.5]$, which were shown in~\cite{inpainting} to improve the quality of generated (recovered) images.

During the linear evaluation, we only preserve the center-cropping data augmentation which follows the practice of~\cite{chen2020simple}. Before updating weights, we accumulate the gradient until the number of batches reaches $4\rm{,}096$. The warmup mechanism is also adopted for $10$ epochs during training. We set the balancing coefficient, $\lambda$, between $\mathcal{L}_\mathrm{D}$ and $\mathcal{L}_\mathrm{G}$ to be $0.1$ as default.

\setcounter{table}{7}
\begin{table}[]
\vspace{0.05in}
\fontsize{9.0}{10.5}\selectfont
\centering
\setlength{\tabcolsep}{3mm}
\begin{tabular}{c|c|c}
\toprule
dim.  &comp. ratio   & Acc. (\%) \\
\midrule
1024    &147  &71.8 \\ 
512     &294  &71.9 \\
256     &588  &71.7 \\
\bottomrule
\end{tabular}
\caption{Results of different settings of the bottleneck vector, which also corresponds to different compression ratios. The accuracy comes from linear evaluation.}
\label{tab:comp_rate}
\end{table}

\begin{table}[]
\vspace{0.05in}
\fontsize{9.0}{10.5}\selectfont
\centering
\setlength{\tabcolsep}{2mm}
\begin{tabular}{c|c|c|c}
\toprule
ID &channel  &param. (M)   & Acc. (\%) \\
\midrule
1   &[1024-512-256-128-64-3]    &21.8  &72.0 \\ 
2   &[512-256-128-64-32-3]     &11.3  &71.9 \\
3   &[256-128-64-32-16-3]     &8.2  &71.7 \\
\bottomrule
\end{tabular}
\caption{Results of using different decoder networks. The channel item in the table denotes the initial channel number, which decreases by a factor of $2$ in each layer. The corresponding spatial resolution at these layers are $7\times7$, $14\times14$, $28\times28$, $56\times56$, $112\times112$, and $224\times224$, respectively. The number of parameters is only for the decoder. The accuracy comes from linear evaluation.}
\label{tab:decoder}
\end{table}

\begin{table}[]
\vspace{0.05in}
\fontsize{9.0}{10.5}\selectfont
\centering
\setlength{\tabcolsep}{2.0mm}
\begin{tabular}{x{28}|x{28}|x{32}|x{42}|x{34}}
\toprule
decoder  &MSE  &evaluator  &multi-level  & Acc. (\%) \\
\midrule
training  &\cmark  &\cmark  &\cmark  &71.9 \\
\midrule
frozen    &\cmark  &\cmark  &\cmark  &70.5 \\ 
training  &\xmark  &\cmark  &\cmark  &71.6  \\ 
training  &\cmark  &\xmark  &--  &67.2  \\ 
training  &\cmark  &\cmark  &\xmark  &71.6  \\ 
\midrule
frozen    &\xmark  &\xmark  &\xmark  &66.8 \\
\bottomrule
\end{tabular}
\caption{Effect of different modules. The training (or frozen) for decoder represents we update (or freeze) the weights of decoder, and \cmark (or \xmark) for MSE, evaluator, and multi-level represent we use them (or not). If multi-level is not used, we only compute the distance between the $1\times1$ features as part of the loss function. The first and last row correspond to the complete framework (SaGe) and the baseline approach (BYOL, in which there is actually no decoder at all). The accuracy comes from linear evaluation.}
\label{tab:modules}
\end{table}

\section{Ablation on Different Modules}

\textbf{\textit{This part complements the contents described in Section~4.3 of the main article.}} We follow the setting of Section~4.3 to pre-train all models for $100$ epochs and report the linear test accuracy.

\begin{figure*}
\centering
\includegraphics[width=17.6cm]{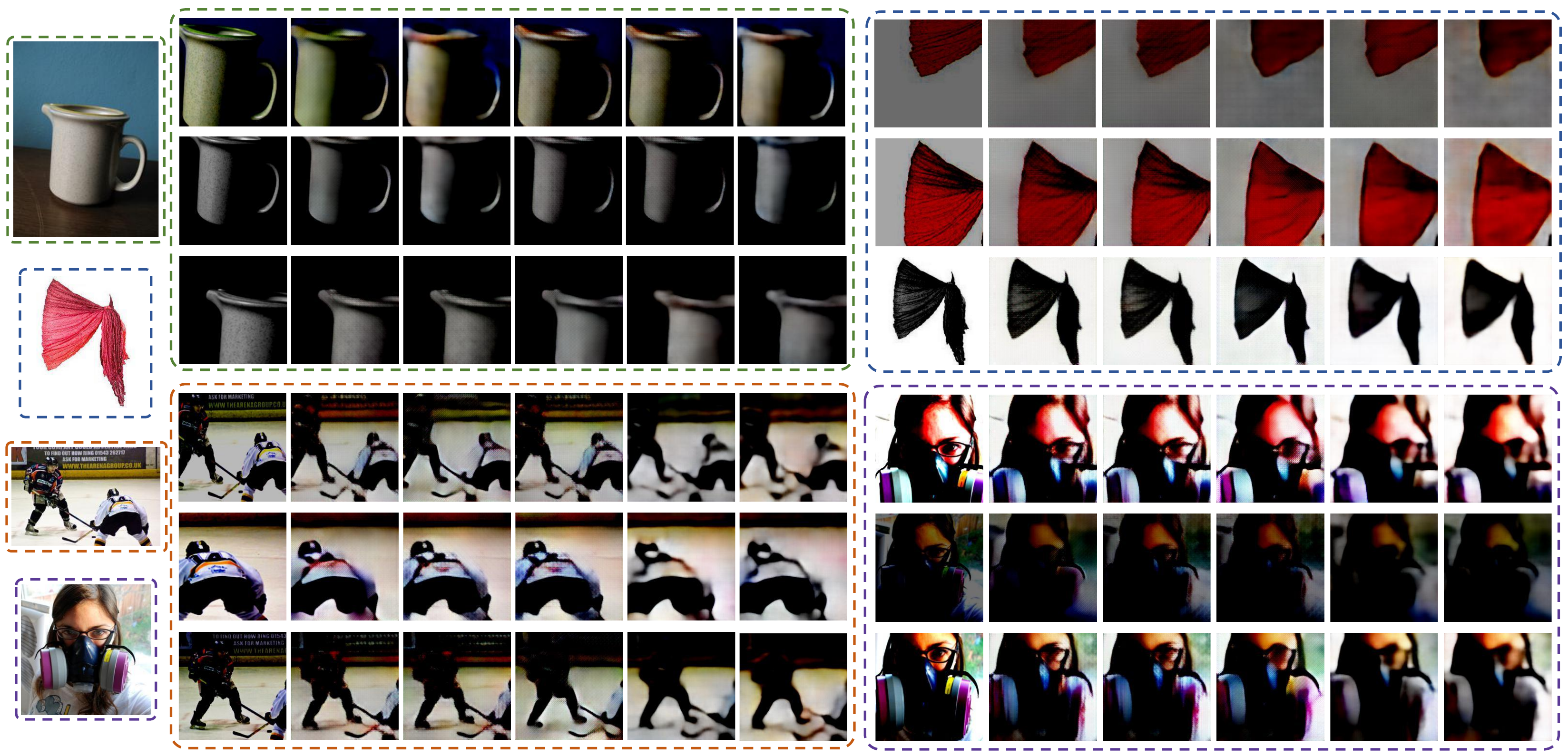}
\caption{Visualization of generated images under different settings. The very first column shows four examples from ImageNet-1K, each of which corresponds to a block (bounded by the same color) in the right-hand side. From each original image, we extract three views and they are shown in the $1^\mathrm{st}$ column of each block. The $2\mathrm{nd}$--$4^\mathrm{th}$ columns correspond to three settings of bottleneck dimensionality, namely, $1\rm{,}024$, $512$, and $256$, respectively. The $5^\mathrm{th}$ column shows the result when multi-level features are not used in the evaluator, and the last ($6^\mathrm{th}$) column shows that an average pooing, rather than a $7\times7$ convolution, is applied to obtain the bottleneck vector.}
\label{fig:supp_visualization}
\end{figure*}

\textbf{First}, we investigate the dimensionality of the $1\times1$ feature vector that bridges the encoder and decoder. We name it the bottleneck representation as it is the most compact vector that propagates the input information to the decoder for image recovery. We test three settings, with the dimensionality of bottleneck representation being $1\rm{,}024$, $512$, and $256$, respectively. From the viewpoint of data compression, we can approximately compute the compression ratio of these settings being $224\times224\times3$ (the image pixels) divided by the bottleneck dimensionality\footnote{we can quantify each entry of the bottleneck vector into a $8$-bit floating point number which merely impacts the quality of image recovery.} in Table~\ref{tab:comp_rate}, we show the effect of different settings. SaGe is generally robust to the change of settings, yet a proper bottleneck dimensionality (\textit{e.g.}, $512$) produces slightly better accuracy. We show the generated images under different compression ratios in Figure~\ref{fig:supp_visualization}.

\textbf{Next}, we investigate the architecture of the decoder by changing the numbers of channels in each layer. Results are shown in Table~\ref{tab:decoder}. Though the decoder serves as an auxiliary network (\textit{i.e.}, it is discarded when the encoder is transferred to the downstream tasks), increasing its size and thus the number of parameters still benefits the test accuracy. The default setting used in the main article (\# 2) achieves a good tradeoff between the training complexity and performance. In comparison, a larger decoder with almost doubled parameters achieves slightly better accuracy, while a smaller decoder with $30\%$ fewer parameters works worse.

\textbf{Lastly}, we analyze the effects of different modules in our framework and validate the effectiveness of our design. The ablation options include freezing the decoder (\textit{i.e.}, not optimizing it during the entire pre-training stage), not using the MSE loss term ($\mathcal{L}_\mathrm{MSE}$), not using the higher-level features of the evaluator (\textit{i.e.}, only using $\mathcal{L}_\mathrm{MSE}$), and not using multi-level features for evaluator (\textit{i.e.}, only using $\mathcal{L}_\mathrm{MSE}$ and $\mathcal{L}_{1\times1}$). As shown in Table~\ref{tab:modules}, each ablation contributes accuracy drop of different extents.


\begin{table*}[]
\renewcommand\arraystretch{1.0}
\setlength{\tabcolsep}{2.0mm}
\fontsize{9.5}{11.5}\selectfont
\centering
\begin{tabular}{l|ccc|ccc|ccc|ccc}
\toprule
\multirow{3}{*}{Method} & \multicolumn{6}{c|}{Mask R-CNN, R50-FPN, detection}  &\multicolumn{6}{c}{Mask R-CNN, R50-FPN,  segmentation} \\ \cline{2-13}
&\multicolumn{3}{c|}{1$\times$ schedule}  &\multicolumn{3}{c|}{2$\times$ schedule}  &\multicolumn{3}{c|}{1$\times$ schedule}  &\multicolumn{3}{c}{2$\times$ schedule} \\ \cline{2-13}
& AP$^{\mathrm{bb}}$  & AP$^{\mathrm{bb}}_{\mathrm{50}}$ & AP$^{\mathrm{bb}}_{\mathrm{75}}$  & AP$^{\mathrm{bb}}$  & AP$^{\mathrm{bb}}_{\mathrm{50}}$ & AP$^{\mathrm{bb}}_{\mathrm{75}}$   &AP$^{\mathrm{mk}}$  &AP$^{\mathrm{mk}}_{\mathrm{50}}$  &AP$^{\mathrm{mk}}_{\mathrm{75}}$   &AP$^{\mathrm{mk}}$  &AP$^{\mathrm{mk}}_{\mathrm{50}}$  &AP$^{\mathrm{mk}}_{\mathrm{75}}$\\ \midrule

Supervised    &38.9  &59.6  &42.7    &38.9  &59.6  &42.0   &35.4  &56.5   &38.1     &35.4  &56.5  &38.1  \\ \midrule
MoCo v1~\cite{he2020momentum} &38.5 &58.9  &42.0    & 40.8 &61.6 &44.7   &35.1  &55.9  &37.7  &36.9 &58.4 &39.7 \\
MoCo v2 ~\cite{chen2020improved} &39.2  &59.9  &42.7  &41.6  &62.1 &45.6  &35.7 &56.8  &38.1    &37.7  &59.3  &40.6   \\
HCL~\cite{HCL}  &40.0  & 60.6  &43.8  &41.8  & 62.4  &45.7   &36.4  &57.6  &39.1  &37.8  &59.5  & 40.8 \\
DenseCL~\cite{wang2021dense}  &40.3 &59.9 &44.3 &-- &-- &--    &36.4 &57.0 &39.2 &-- &-- &-- \\ 
self-EMD~\cite{liu2020self}  &40.0   &60.4  &44.0  &-- &-- &--  &-- &--	&-- &-- &-- &--\\
ORL~\cite{ORL} &40.3  &60.2  &44.4  &-- &-- &--  &36.3  &57.3  & 38.9  &-- &-- &-- \\ 
\midrule

\textbf{SaGe} (300 epochs) & 40.2  & 61.7  & 43.8  &41.9  &63.0  &45.7  &36.7  &58.3  &39.2  &38.0  &59.9  &40.8 \\
\textbf{SaGe} (800 epochs) & 40.8  & 62.4  & 44.8  &42.3  &63.6  &46.1  &37.2  &59.0  &40.1  &38.3  &60.4  &41.0\\ 
\bottomrule
\end{tabular}
\caption{Object detection and instance segmentation APs (\%) on the MS-COCO dataset. The results of $1\times$ training schedule have been reported in the main article and we supplement the results of $2\times$ training schedule for comparison.}
\label{tab:supp_coco}
\end{table*}

\section{More Results on COCO}

\textbf{\textit{This part complements the MS-COCO experiments shown in Section~4.2 of the main article.}}

We perform object detection and instance segmentation results on the MS-COCO dataset with the $2\times$ schedule, and report the results together with that of $1\times$ schedule in Table~\ref{tab:supp_coco}. As in other downstream transfer tasks, SaGe still shows favorable performance in the $2\times$ schedule.

\end{document}